\documentclass[letterpaper, 10 pt, conference]{ieeeconf}  

\IEEEoverridecommandlockouts                              

\usepackage{amsmath} 
\usepackage{amssymb}  
\usepackage{graphicx}
\usepackage{xcolor}
\usepackage{color}
\usepackage{hyperref}
\usepackage{upgreek}

\title{\LARGE \bf
Koopman Operator Based Linear Model Predictive Control for Quadruped Trotting
}

\author{Chun-Ming Yang and Pranav A. Bhounsule
\thanks{Dept. of Mechanical and Industrial Engineering, University of Illinois at Chicago, 
       842 W Taylor St, Chicago, IL 60607, USA. Email:
       {\tt\small jyang241@uic.edu}, {\tt\small pranav@uic.edu}
       The work was supported by NSF grant 2128568.}%
}

\begin{document}

\maketitle
\thispagestyle{empty}
\pagestyle{empty}

\begin{abstract}
Online optimal control of quadruped robots would enable them to adapt to varying inputs and changing conditions in real time. A common way of achieving this is linear model predictive control (LMPC), where a quadratic programming (QP) problem is formulated over a finite horizon with a quadratic cost and linear constraints obtained by linearizing the equations of motion and solved on the fly. However, the model linearization may lead to model inaccuracies. In this paper, we use the Koopman operator to create a linear model of the quadrupedal system in high dimensional space which preserves the nonlinearity of the equations of motion. Then using LMPC, we demonstrate high fidelity tracking and disturbance rejection on a quadrupedal robot. This is the first work that uses the Koopman operator theory for LMPC of quadrupedal locomotion. 
\end{abstract}

\section{Introduction}

Quadruped robots, due to their low center of gravity, wide base of support, and ability to move over uneven terrain and obstacles using limited footholds, provide a viable means of using mobile robots in applications such as first responders, industrial workers, and helpers at home. In this regard, the low-level or joint-level control of these robots is of paramount importance. Current model-based approaches restrict to linearized models while model-free approaches use large sample size. This paper addresses the problem of generating low-level control of quadrupeds in a generalizable, sample-efficient manner without resorting to model linearization using the Koopman operator theory.

The traditional approach of performing low-level control is to create a parametric controller and then fine tune gains either in simulation and/or hardware \cite{robotics12020035}. However, this approach may not scale to different quadrupeds and/or scenarios because it involves non-intuitive hand tuning. This can be overcome by formulating and solving a trajectory optimization (TO) problem offline by tuning a cost function subject to state and control constraints. The result of TO is a reference trajectory and/or open loop torque profile. To implement TO on hardware, one needs an additional feedback control layer to ensure high-level tracking in the presence of disturbances and noise \cite{bhounsule2013recordwalk}.

Model predictive control (MPC) is similar to TO in that it solves an optimal control problem. However, there are two major differences: one, it solves the TO online over a finite horizon to make it computationally efficient, and two, it repeatedly updates the control using sensor measurements, thus providing accurate tracking even in the presence of disturbances and noise. MPC for quadruped control involves using a physics model, such as single rigid body (SRB), an idealization, for planning the reaction forces and then mapping those forces to joint torques using the Jacobian \cite{kim2019highly}. Some of the nonlinearities in SRB may be ignored or linearized to avoid formulating a nonlinear MPC problem, which is challenging to solve in real time \cite{ding2021representation}. The limitations of SRB, such as model mismatch, can be avoided by using a whole-body model for MPC, but one needs an engineering feat to make the online computation fast enough for real-time control \cite{neunert2018whole}. 
\begin{figure} [tbp]
\centering
\includegraphics[scale=0.25]{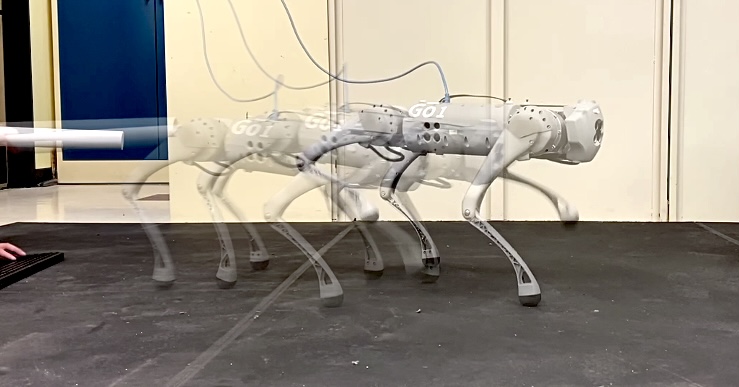}
\caption{Push recovery using Koopman operator-based linear model predictive control. Video summary and experimental results:    \url{https://youtu.be/1JziT5hljjM} }
\label{fig:false-color}
\vspace{-0.5cm}
\end{figure}
Deep Reinforcement learning (DRL) is an model-free method that solves the optimization problem over the entire state space in an offline manner to learn a neural network that maps the control to the states. To enable seamless transfer to hardware, one could learn trajectories which are then implemented using feedback control on hardware \cite{feng2022genloco}. Another approach is to use dynamic randomization, where torque profiles are generated for a wide variety of model and environmental parameters such that when the controller is transferred to hardware, it is able to cope with the sim-to-real gap \cite{chen2023learning}. However, DRL is not sample efficient. To achieve sample efficiency, one could use model-based TO to plan reference motion and model-free RL to track the reference motion \cite{jenelten2024dtc}. 

The Koopman operator takes a nonlinear model $\mathbf{x}_{t+1}= \mathbf{f}(\mathbf{x}_{t}, \mathbf{u}_{t} )$ and converts it into a linear model in high dimensional space: $\mathbf{\Pi}(\mathbf{x}_{t+1})=\mathbf{A}\mathbf{\Pi}(\mathbf{x}_{t})+\mathbf{B}\mathbf{u}_{t}$, where ${\bf A}, {\bf B}$ are constant matrices and ${\bf \Pi({\bf x})}$ is a non-linear function of ${\bf x}$. The original method was conceptualized for an uncontrolled system in 1931 by Koopman \cite{koopman1931hamiltonian}, but only recently tools have been devised to model controlled systems \cite{williams2015data}. The applications of the Koopman operator are currently limited to a few simple smooth systems, such as quadcopters \cite{narayanan2023se}, underwater vehicles \cite{rahmani2024enhanced}, autonomous cars \cite{kim2001model}, two-link planar manipulators \cite{shi2021acd}, and soft robot manipulators \cite{bruder2020data}. 

In this paper, we use the Koopman operator to create a relatively low degree of freedom, a linear model of a quadruped, and use Linear Model Predictive Control (LMPC) to demonstrate slip and push recovery, reference tracking in translation, and rotation. The main novelty of this work is that it is the first demonstration of the use of Koopman operator theory for modeling and controlling a quadrupedal robot. The method is generalizable to other quadruped system because we use SRB, a well known idealization of many quadruped robots, for system identification. Finally, hardware demonstration shows the efficacy of the approach to do real-time optimal control.

\begin{figure} [tbp]
\includegraphics[scale=1.8]{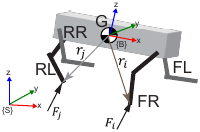}
\caption{Single rigid body model}
\label{fig:SRB}
\vspace{-0.5cm}
\end{figure}

\section{Methods}
\subsection{Single Rigid Body Model (SRB)}
The quadruped system shown in Fig.~\ref{fig:SRB} is modeled using SRB model \cite{di2018dynamic} 
%
%
%
 \begin{align} 
 \mathbf{\dot{\Theta}} &= \mathbf{A}_{\Omega}^{-1}(\theta, \psi)\mathbf{\Omega} \label{eqn:SRB1} \\   
 \mathbf{\ddot{p}} &= \frac{ \mathbf{R} (\mathbf{F}_{i} + \mathbf{F}_{j})}{m} - \mathbf{g}  \label{eqn:SRB2} \\
  \mathbf{I \dot{\Omega}} &= 
   \mathbf{\Omega} \times (\mathbf{I {\Omega}}) +
 \mathbf{r}_{i} \times\mathbf{F}_{i} + \mathbf{r}_{j} \times\mathbf{F}_{j} \label{eqn:SRB3} 
\end{align}
where 
$\mathbf{p} \in \mathbb{R}^{3}$  and $ \mathbf{\dot{p}} \in  \mathbb{R}^{3}$  are the SRB center of mass (CoM) position and velocity both in the world frame respectively; $ \mathbf{\Theta}  =[\phi, \theta, \psi]^\top$, $ \mathbf{\Omega} \in  \mathbb{R}^{3}$ are orientation expressed by X-Y-Z Euler angles and angular velocities in body frame respectively. The leg ground reaction force for the foot $i$ is $ \mathbf{F}_{i} \in \mathbb{R}^{3} $ and the distance from the foot to the center of mass is $ \mathbf{r}_{i}  \in \mathbb{R}^{3} $;
$ \mathbf{A}_{\Omega}(\theta, \psi) =  [c_{\psi}/c_{\theta}, -s_{\psi}/c_{\theta}, 0; s_{\psi}, c_{\psi}, 0; -c_{\psi}s_{\theta}/c_{\theta}, s_{\psi}s_{\theta}/c_{\theta}, 1]\in  \mathbb{R}^{3 \times 3}$ is the matrix transforms body angular to Euler angle rate, where $c_{A}=\cos A,s_{A} = \sin A$; $\mathbf{R} \in \mathbb{R}^{3 \times 3}$ is the rotation matrix mapping vector from body frame to global frame; $m$ is the mass of the robot; $ \mathbf{g} \in  \mathbb{R}^{3}$ is the gravity vector in the world frame; $\mathbf{I} \in \mathbb{R}^{3}  $ is the inertia matrix of the torso in the body frame. Note that Eqns.~\ref{eqn:SRB1}-\ref{eqn:SRB3} are nonlinear and they can be compactly written as: $\dot{{\bf x}} = {\bf f}({\bf x}) + {\bf g}({\bf x}) {\bf u}$,  where $\mathbf{x}=[\mathbf{p},    \mathbf{\Theta},      \mathbf{\dot{p}},   \mathbf{\Omega}   ]^\top $ and ${\bf u} = [{\bf F}_i,{\bf F}_j]^\top$.


\subsection{Koopman Operator Theory}

For a given nonlinear system $ \mathbf{x}_{t+1}= \mathbf{f}(\mathbf{x}_{t}, \mathbf{u}_{t} ) ; \ \mathbf{x} \in \mathcal{X} \subseteq \mathbb{R}^n; \ \mathbf{u} \in \mathcal{U} \subseteq \mathbb{R}^m; \ \mathbf{f} : \mathcal{X} \to \mathcal{X} $, a set of nonlinear observable functions  $ \mathbf{\Pi}(\mathbf{x}) $ exists such that the evolution of the system along these observables is characterized by linear dynamics governed by an infinite dimension operator $ \mathbf{\mathcal{K}} $, known as Koopman operator
\begin{align}
    [\mathbf{\mathcal{K}}\mathbf{\Pi}](\mathbf{x},\mathbf{u})= \mathbf{\Pi} \circ \mathbf{f}( \mathbf{x}, \mathbf{u} ) 
\end{align}
A finite-dimensional approximation of $ \mathbf{\mathcal{K}}  $, denoted as $ \mathbf{K} = [\mathbf{A}, \mathbf{B}];  \mathbf{A} \in \mathbb{R}^{N \times N};  \mathbf{B} \in \mathbb{R}^{N \times m} $, is derived by employing the Extended Dynamic Mode Decomposition (EDMD) approach \cite{williams2015data}, which projects $ \mathbf{\mathcal{K}} $ onto a subspace of  observable functions via least squares regression. The finite dimension approximated operator $ \mathbf{K} $ can be used to represent the linear evolution of observable functions as
%
\begin{align}
    \mathbf{\Pi}(\mathbf{x}_{t+1}) = \begin{bmatrix} \mathbf{A}& \mathbf{B} \end{bmatrix} \begin{bmatrix} \mathbf{\Pi}(\mathbf{x}_{t}) \\ \mathbf{u}_{t} \end{bmatrix} = \mathbf{K}\mathbf{\hat{\Pi}}(\mathbf{x}_{t}, \mathbf{u}_{t}) \label{eqn:KP}
\end{align}
where $\mathbf{\Pi}(\mathbf{x}_{t}) = [\pi_{1}(\mathbf{x}_{t}),...,\pi_{N}(\mathbf{x}_{t})]^T \in \mathbb{R}^{N}$ is the dictionary of observable functions. By utilizing $M$ snapshots of the system states and control inputs in lifted space formed by the basis function $ \pi_i(\mathbf{x}) $ in the dictionary, we obtain the approximated operator $ \mathbf{K} $. The paired dataset $ \mathbf{X}=[\mathbf{x}_{1},\mathbf{x}_{2},...,\mathbf{x}_{M-1}]$ and $ \mathbf{Y} = [\mathbf{x}_{2},\mathbf{x}_{3},...,\mathbf{x}_{M}] $ can be obtained by perturbed nonlinear dynamics $ \mathbf{x}_{t+1}=\mathbf{f}(\mathbf{x}_{t}, \mathbf{u}_{t})  $ with a given control sequence $ \mathbf{U}=[\mathbf{u}_{1},\mathbf{u}_{2},...,\mathbf{u}_{M-1}] $, the approximation of the Koopman operator then can be obtained by solving the least square regression such that \cite{li2017extended}
%
\begin{align}
\mathbf{K} = \arg\min_{\mathbf{K}} \|  \mathbf{\Pi}(\mathbf{Y}) - \mathbf{K} 
 \mathbf{\hat{\Pi}}(\mathbf{X, U})   \|^{2}
\end{align}
By constructing $ \mathbf{G}_{1} $ and $ \mathbf{G}_{2} $, an analytical solution for $\mathbf{K}$ may be computed
\begin{align}
    \mathbf{G}_{1} &= \frac{1}{M} \sum_{i=1}^{M} \mathbf{\Pi}(\mathbf{y}_i) \hat{\mathbf{\Pi}}(\mathbf{x}_i, \mathbf{u}_{i})^\top \\
\mathbf{G}_{2} &= \frac{1}{M} \sum_{i=1}^{M} \mathbf{\Pi}(\mathbf{x}_i) \hat{\mathbf{\Pi}}(\mathbf{x}_i, \mathbf{u}_{i})^\top \\
\mathbf{K} &=\mathbf{G}_{1}\mathbf{G}_{2}^{-1}
\end{align}

\subsection{Koopman Operator-Based Modeling} \label{sec:KP_model}
The SRB model equations (see Eqn.~\ref{eqn:SRB1}-\ref{eqn:SRB3}) are nonlinear. Our goal is to compute a linear model using Koopman operator theory.

We identify a set of Koopman observer functions  $ \mathbf{\Pi}(\mathbf{x}): \mathbb{R}^{n} \rightarrow \mathbb{R}^{N } $ that can evolve linearly in the lifted observable space with the finite-dimensional Koopman operator $ \mathbf{K}=[\mathbf{A}, \mathbf{B}] $, such that the dynamics are approximated by Eqn.~\ref{eqn:KP}. 
To perform the EDMD to find the linear Koopman predictor $ \mathbf{A} \in \mathbb{R}^{N \times N}$ and  $\mathbf{B} \in \mathbb{R}^{N \times m} $ , a set of physics informed observable functions \cite{narayanan2023se} $ \bar{\mathbf{\Pi}} = [\underline{\mathbf{R}\mathbf{\Omega}}, \underline{\mathbf{R}\mathbf{\Omega}^{2}}, ..., \underline{\mathbf{R}\mathbf{\Omega}^{p}}]^\top $ is selected to form the lifted state space, 
 where the operator $ \underline{(\cdot)} : \mathbb{R}^{l \times l} \rightarrow \mathbb{R}^{l^2 }$ maps the matrix into a vector by concatenating the columns inside the matrix, then the linear SRB states can be augmented as
\begin{align}
    \mathbf{\Pi} = [1, \mathbf{p},    \mathbf{\Theta},      \mathbf{\dot{p}},   \mathbf{\Omega}, \bar{\mathbf{\Pi}}]^\top \in   \mathbb{R}^{13+9p}
\end{align}
Note that one of the observable is $1$ here is needed to put the constant terms such as the gravity term. We also include the state ${\bf x}=[{\bf p}, {\bf \Theta}, {\bf {\dot p}},{\bf \Omega}]^\top$ so we can recover it for use in the LMPC discussed in the next section.





\begin{figure*}[tbp]
  \centering
   \includegraphics[scale=0.7]{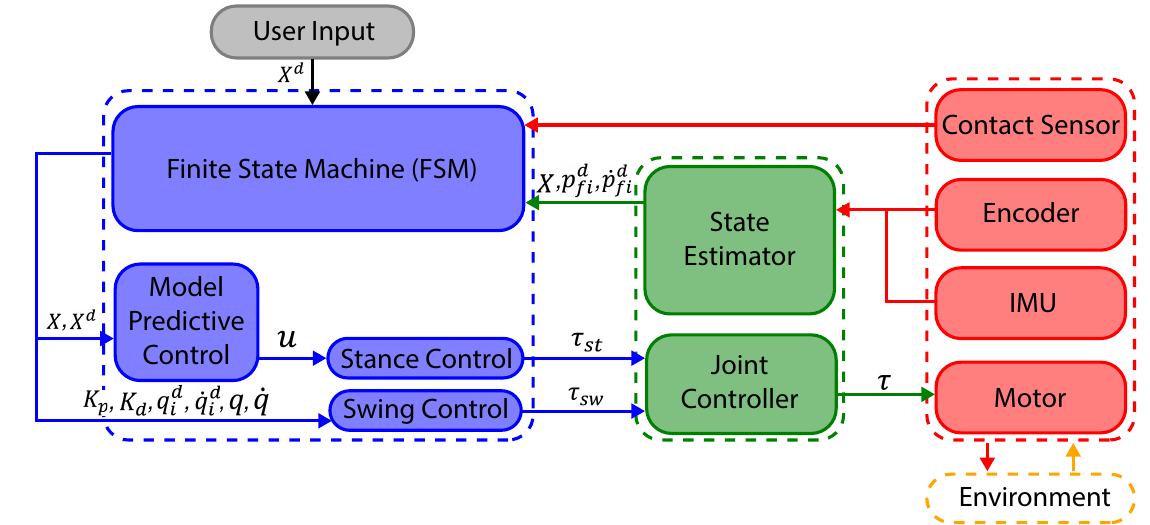}
 \caption{Control diagram for implementation on the quadruped in simulation and hardware}
\label{fig:control_diagram}
\end{figure*}



\subsection{Koopman Operator-Based Model Predictive Control} \label{sec:KP_MPC}
We formulate a LMPC using the Koopman operator model as follows
%
%
\begin{align} 
&\min_{\mathbf{u}}\sum_{i=0}^{k-1}\left\| \mathbf{x}_{t+i}-\mathbf{x}^{d}_{t} \right\|_{\mathbf{Q}_{i}} + \left\| \mathbf{u}_{t+i} \right\|_{\mathbf{R}_{i}} \label{eqn:MPC_cost} \\
 \text{s.t.} &\quad \mathbf{\Pi}_{t+i} = \mathbf{A}\mathbf{\Pi}_{i} + \mathbf{B}\mathbf{u}_{i}, \: i=0,1...k-1  \label{eqn:MPC_model1} \\
 &\quad \mathbf{x}_{i}= {\bf C}_x\mathbf{\Pi}(\mathbf{x}_{i}), \:{\bf u}_{\mbox{\scriptsize min}} \leq {\bf u}_i \leq {\bf u}_{\mbox{\scriptsize max}}   \label{eqn:MPC_control3} 
\end{align}
where ${\bf C}_x \in \mathbb{R}^{n \times N} $ is selection matrix that pulls out the state ${\bf x}$ from ${\bf \Pi(\mathbf{x})}$; 
$ \mathbf{Q}_{i} \in \mathbb{R}^{n \times n} $ and $ \mathbf{R}_{i} \in  \mathbb{R}^{m \times m} $ are user-chosen diagonal positive definite matrices.



For a given initial state ${\bf x}_0 \in \mathbb{R}^{n} $, using Eqn.~\ref{eqn:MPC_model1} and \ref{eqn:MPC_control3} recursively for $k$ time steps, we obtain 
\begin{align}
    \mathbf{X}_{qp} = \mathbf{A}_{qp}\mathbf{x}_{0} + \mathbf{B}_{qp}\mathbf{U}_{qp}
\end{align}
where $\mathbf{X}_{qp} \in  \mathbb{R}^{n \times k} $ and $ \mathbf{U}_{qp} \in \mathbb{R}^{m \times k} $  are the concatenated state and control from $1,2,...,k$
%
The cost function can then be rewritten as
\begin{align}
    &\min_{\mathbf{U}_{qp}}\left\| \mathbf{A}_{qp}\mathbf{x}_{0} + \mathbf{B}_{qp}\mathbf{U}_{qp}-\mathbf{X}^{d}_{qp} \right\|_{\mathbf{Q}_{qp}} + \left\| \mathbf{U}_{qp} \right\|_{\mathbf{R}_{qp}}
\end{align}
where $ \mathbf{X}^{d}_{qp} \in    \mathbb{R}^{n \times k}$ is the concatenated reference trajectories; \(\mathbf{Q}_{qp} \in \mathbb{R}^{nk \times nk} \) and \(\mathbf{R}_{qp} \in  \mathbb{R}^{mk \times mk} \) are user-chosen diagonal positive weight matrix.

The QP can now be written as
\begin{align}
    &\min_{\mathbf{U}_{qp}} \quad \frac{1}{2}\mathbf{U}_{qp}^\top \mathbf{H} \mathbf{U}_{qp} + \mathbf{P} \mathbf{U}_{qp} \\
    \text{s.t.} &\quad \mathbf{\underline{c}}  \leq \mathbf{C}\mathbf{U}_{qp} \leq  \mathbf{\overline{c}} 
\end{align}
where $\mathbf{H} = 2(\mathbf{B}_{qp}^\top \mathbf{Q}_{qp} \mathbf{B}_{qp}+ \mathbf{R}_{qp})$, $\mathbf{P} = 2(\mathbf{x}_{0}^\top \mathbf{A}_{qp}^\top \mathbf{Q}_{qp} \mathbf{B}_{qp} - \mathbf{X}_{qp}^{d\mathstrut\top}  \mathbf{Q}_{qp} \mathbf{B}_{qp})$ and $\mathbf{C} \in \mathbb{R}^{mk \times mk}, \mathbf{\underline{c}} \in \mathbb{R}^{mk}, \mathbf{\overline{c}} \in \mathbb{R}^{mk}$ denoted the inequality constraints of control input.

\subsection{Controller Implementation in Simulation and Hardware}
The controller is implemented on a model of Unitree Go1 quadruped in MuJoCo version 2.0.0 \cite{todorov2012mujoco} using Ubuntu 20.04 on an Intel Core i7 and on hardware. 

Each leg of the Go1 robot comprises two $0.2$ m links with three degrees of freedom. The torso weighs $4.75$ kg, and each leg weighs $2$ kg. The torso has an IMU providing orientation, angular speed, and linear acceleration. Each joint has an absolute encoder providing joint angles. The joint torque limits are $\pm33.5$ Nm. An onboard Raspberry Pi 4 processes the IMU and encoder data and sends torque, position, and velocity references along with their gains to the motor controller at $1000$ Hz. The motor controller operates at $10$ kHz to ensure high-fidelity torque tracking. Figure~\ref{fig:control_diagram} gives an overview of the Koopman operator-based LMPC in simulation and hardware and  details follow 

\subsubsection{Finite State Machine}
A finite state machine is used to manage the control logic, sending commands  to leg controllers for Front Right (FR), Front Left (FL), Rear Right (RR), and Rear Left (RL) legs. In a trot gait, FR-RL and FL-RR legs move in pairs, transitioning between swing and stance phases every $0.2$ sec. If a swing leg collides with an obstacle, detected by a contact sensor, the swing foot's position is held constant.

\subsubsection{Swing Leg Controller}
The role of the swing leg controller is to track the reference position of the swinging foot using joint torques on the swing leg joints. Given the swing foot reference position and velocity, $\mathbf{p}_{fi}^{d},\mathbf{\dot{p}}_{fi}^{d} $,  an analytical inverse (see \cite{robotics12020035} Sec. 3.5) is used to compute the corresponding joint reference position and velocity, $ \mathbf{q}_{i}^{d} ,\mathbf{\dot{q}}_{i}^{d} $.  The following simple proportional-derivative $ \mathbf{K}_{p},\mathbf{K}_{d} $ controller is used to compute the joint torques $ \mathbf{\tau}_{i} $ for leg $i$ at $1$ kHz.
\begin{align}
    \mathbf{\tau}_{i} = -\mathbf{K}_{p}(\mathbf{q}_{i} - \mathbf{q}_{i}^{d})  - \mathbf{K}_{d}(\mathbf{\dot{q}}_{i} - \mathbf{\dot{q}}_{i}^{d} )
\end{align}

\subsubsection{Linear Model Predictive Control}
The LMPC uses the Koopman operator-based model and estimated torso states to compute desired ground reaction forces for the stance legs (see Sec.~\ref{sec:KP_MPC}). The planning horizon for the model predictive control is $6$ ms or $166.67$ Hz while the update horizon is $5$ ms or $200$ Hz. The MPC is solved online using qpSWIFT \cite{pandala2019qpswift} in about $3$ ms. 

\subsubsection{Stance Leg Controller}
The stance leg controller module uses the desired ground reaction forces $ \mathbf{f}_{i} $ to estimate the joint torques $ \mathbf{\tau}_{i} $ using the Jacobian $ \mathbf{J}_{i} $ of the $i$ stance leg. \begin{align}
    \mathbf{\tau}_{i} = \mathbf{J}_{i}^\top{} \mathbf{f}_{i}
\end{align}

\subsubsection{State Estimation}
The state estimation module estimates the linear velocity of the torso by fusing the acceleration data from the IMU and the joint encoder data using an Extended Kalman Filter (EKF). The EKF uses the integrated acceleration as the process model and the estimated linear velocity using the Jacobian of the contact feet and the joint velocity as the measurement model.

\section{Results}



\begin{figure} [tbp]
\includegraphics[scale=0.95]{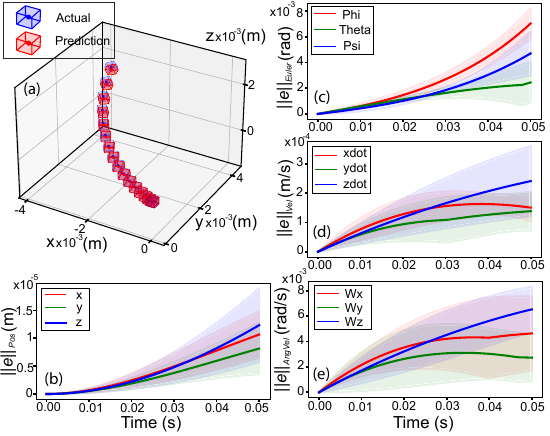}
\caption{Fitting the single rigid body model using the Koopman operator}
\label{fig:koopman_fit}
\end{figure}

\subsection{Koopman Operator Model Fit}

 To generate data for the EDMD, we integrate the SRB Eqns.~\ref{eqn:SRB1}-\ref{eqn:SRB3} with Runge-Kutta of order 4 with a fixed step size of $dt=0.001$ sec from $t=0$ to $t=0.1$ sec. For each roll-out, we use a randomly generated initial conditions at $t=0$ and random control input for every $0.001$ sec. We use a total of $100$ roll-outs to create a training dataset. Then using the $p=4$ observables ${\bf \Pi}({\bf x})$ discussed in Sec.~\ref{sec:KP_model}, we perform the EDMD to obtain a linear model. 
 
 Figure~\ref{fig:koopman_fit} (a) shows fidelity of the fit on one out of fifty testing trajectories. We did the extensive test by generating  $50$ initial conditions and using $50$ random force profiles for a $0.05$ sec roll-out using the SRB model, then using the same paired initial condition and force inputs we generated corresponding predictions using Koopman operator model. The linear model fitting results are evaluated by the error between the actual and predicted trajectories across 50 sets of data  shown in Fig.~\ref{fig:koopman_fit}, where (b) for translation ${\bf p}$, (c) for orientation ${\bf \Theta}$ , (d) for linear velocity ${\bf \dot{p}}$, (e) for angular rates ${\bf \Omega}$. The solid line shows the mean and the bands shows the variance. It can be seen that the errors are within $\pm 10^{-3}$ indicating a sufficiently accurate fit.

\subsection{Koopman Operator LMPC in Simulation}

\begin{figure} [tbp]
\includegraphics[scale=.9]{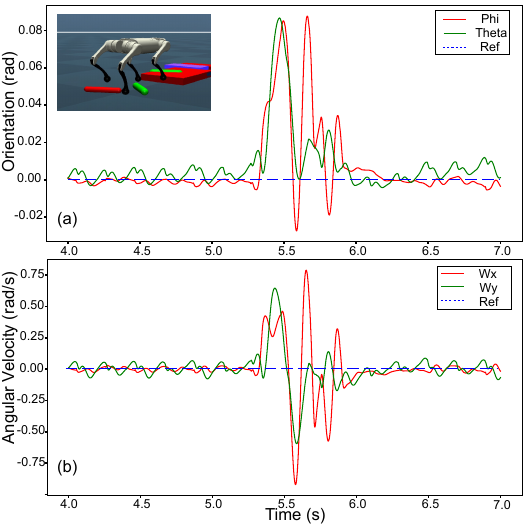}
\caption{Simulation: slip and recovery}
\label{fig:sim_push}
\end{figure}

To check the disturbance rejection, approximately $10$ objects of height in the range of $5-7$ cm are scattered on the ground to intentionally make the robots' feet slip (see Fig.~\ref{fig:sim_push} inset). During the rough terrain simulation experiment, the first significant slip occurs when the robot's rear foot strikes the second object, causing the robot to lose its balance. However, the robot regains stability when the foot makes contact with the flat ground within $1$ sec, recalculating the adequate force for effective disturbance rejection with orientation $ \mathbf{\Theta} $ and angular velocity $ \mathbf{\Omega} $ tracking errors (RMSE) of 0.012 and 0.092 respectively.

\begin{figure} [htbp]
\includegraphics[scale=.9]{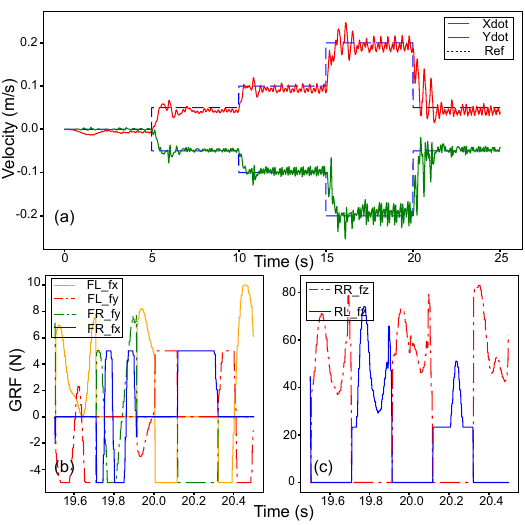}
\includegraphics[scale=.9]{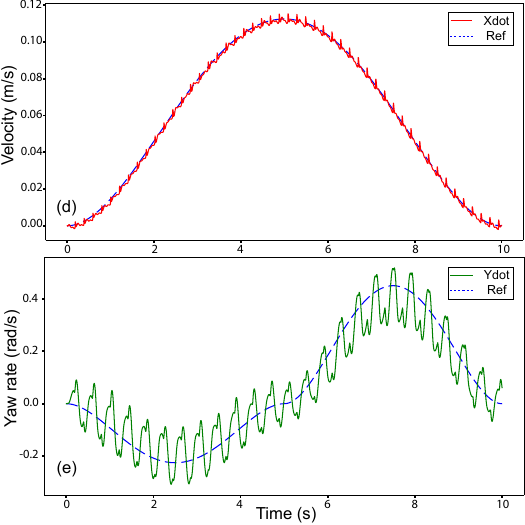}
\caption{Simulation: tracking a forward and lateral speed.}
\label{fig:sim_vx_vy}
\end{figure}

To check the reference tracking ability, we command the robot to follow a combination of reference forward and lateral speed as shown in Fig.~\ref{fig:sim_vx_vy} (a), the RMSE for linear velocity $ \mathbf{\dot{p}} $   is 0.021. Then a combination of forward speed and turning speed as shown in Fig.~\ref{fig:sim_vx_vy} (d) (e) is commanded with resulting tracking RMSE for linear velocity $\mathbf{\dot{p}}$ as 0.008, and for angular velocity $ \mathbf{\Omega} $ as 0.056.




\subsection{Koopman Operator LMPC in Hardware}
\begin{figure}
\includegraphics[scale=.9]{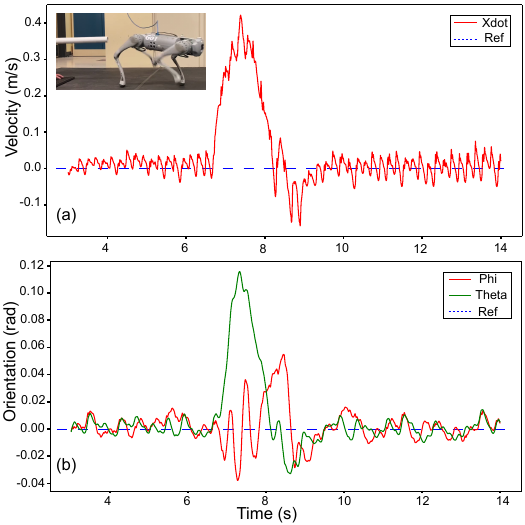}
\caption{Hardware: push and recovery}
\label{fig:hardware_push}
\end{figure}

To evaluate the controller's disturbance rejection capabilities, the robot, while trotting, is subjected to a forceful push that causes its forward velocity to oscillate by up to 0.4 m/s (see Fig.~\ref{fig:hardware_push}). Despite this perturbation, the robot successfully recovers from a pitch deviation of 6.5 degrees within 2 sec with the tracking errors for orientation $ \mathbf{\Theta} $ and angular velocity $ \mathbf{\Omega} $, measured as RMSE, being 0.018 and 0.404, respectively.

\begin{figure}
\includegraphics[scale=.9]{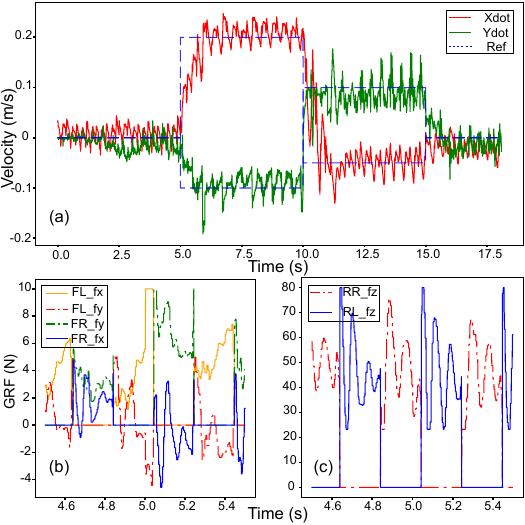}
\includegraphics[scale=.9]{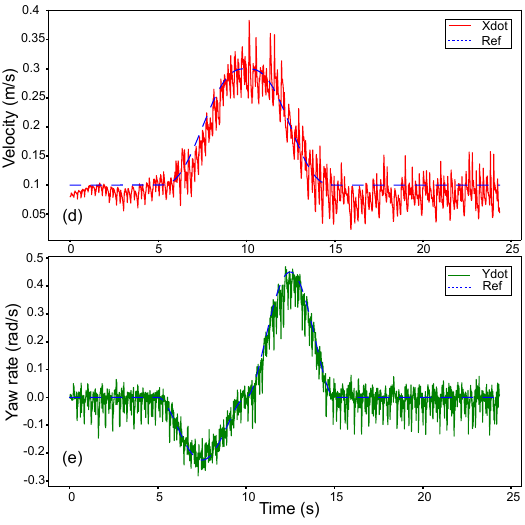}
\caption{Hardware: tracking a fore-back and side-way speed.}
\label{fig:hardware_vx_vy}
\end{figure}

To evaluate the reference tracking capability, we command the robot to follow a combination of forward and lateral reference speeds as shown in Fig.~\ref{fig:hardware_vx_vy} (a). The ground reaction forces in Fig.~\ref{fig:hardware_vx_vy} (b) (c), zoomed in between $4.6-5.4$ sec, demonstrates the controller's ability to manage the steep velocity change from $0$ to $0.2$ m/s for forward and $0$  to $-0.1$ m/s for lateral as commanded by a  step function, with the tracking error of the linear velocity $ \mathbf{\dot{p}} $ quantified by RMSE of $0.026$, lastly, a combination of forward and turning speed is commanded as shown in Fig.~\ref{fig:hardware_vx_vy} (d) (e). The RMSE is $0.048$ for linear velocity $ \mathbf{\dot{p}} $ and $0.415$ for $\mathbf{\Omega}  $ angular velocity.



\section{Discussion, Conclusion, and Future Work}
In this paper, we have used the Koopman operator to create a linear model of the SRB model of a quadruped. This is used with LMPC to control the trot gait of a quadruped in simulation and hardware demonstrating the efficacy of the approach.

We use the analytical model (the SRB model) instead of hardware data for system identification. Often hardware data based data identification is challenging because there may not be enough excitation to tease out the model parameters \cite{jovic2015identification}, but with simulated data it is always possible to generate a richer data set. Although we have used data-based EDMD to identify the linear model, analytical method such as direct encoding may also be used \cite{harry2019dual}.

The SRB model enables us to reduce the dimension of the Koopman operator substantially. The full robot dynamics has $18$ degrees of freedom ($6$ in the torso and $3$ in each of the $4$ legs). However, if we assume the legs are massless, then we only need to model the torso. This is done using the SRB model which has only $6$ degrees of freedom. The result is that we can adequately describe the nonlinear model using only $49$ observables. 

The use of the Koopman operator based linear model leads to a QP over a finite horizon (LMPC) which has an analytical solution if there are no constraints. In our case, we do have control constraints and hence need to solve the problem numerically. However, we are able to solve the LMPC in $3$ ms for a horizon of $6$ ms enabling real-time optimal control. 

 We do see that our linear model remains accurate with a prediction error within $ \pm 10^{-3} $ up to $0.05s$, its accuracy diminishes as time progresses. The SRB model is a control affine model and a bilinear Koopman operator-based model should better describe the dynamics \cite{bruder2021advantages}. But the use of bilinear model leads to a nonlinear model predictive control which is computationally challenging to solve \cite{folkestad2021koopman}. One approach is to linearize the bilinear term at the operator point, which leads to an LMPC \cite{yu2022autonomous}.

In conclusion, the generation of simple linear model with only $49$ observables using the idealized SRB model is found to be adequate to do high-fidelity LMPC of a quadruped in hardware for reference tracking and moderate disturbances. 

Our future work will explore methods to increase the robustness of the quadruped to larger disturbances, increase the range of movement of the robot, and explore multiple gaits such as bounding, pacing, trotting, and transitioning between these gaits. 

\bibliographystyle{IEEEtran}
\bibliography{pranav_bib2} 

\begin{thebibliography}{10}
\providecommand{\url}[1]{#1}
\csname url@samestyle\endcsname
\providecommand{\newblock}{\relax}
\providecommand{\bibinfo}[2]{#2}
\providecommand{\BIBentrySTDinterwordspacing}{\spaceskip=0pt\relax}
\providecommand{\BIBentryALTinterwordstretchfactor}{4}
\providecommand{\BIBentryALTinterwordspacing}{\spaceskip=\fontdimen2\font plus
\BIBentryALTinterwordstretchfactor\fontdimen3\font minus \fontdimen4\font\relax}
\providecommand{\BIBforeignlanguage}[2]{{%
\expandafter\ifx\csname l@#1\endcsname\relax
\typeout{** WARNING: IEEEtran.bst: No hyphenation pattern has been}%
\typeout{** loaded for the language `#1'. Using the pattern for}%
\typeout{** the default language instead.}%
\else
\language=\csname l@#1\endcsname
\fi
#2}}
\providecommand{\BIBdecl}{\relax}
\BIBdecl

\bibitem{robotics12020035}
\BIBentryALTinterwordspacing
P.~A. Bhounsule and C.-M. Yang, ``A simple controller for omnidirectional trotting of quadrupedal robots: Command following and waypoint tracking,'' \emph{Robotics}, vol.~12, no.~2, 2023. [Online]. Available: \url{https://www.mdpi.com/2218-6581/12/2/35}
\BIBentrySTDinterwordspacing

\bibitem{bhounsule2013recordwalk}
P.~A. Bhounsule, J.~Cortell, A.~Grewal, B.~Hendriksen, J.~D. Karssen, C.~Paul, and A.~Ruina, ``Low-bandwidth reflex-based control for lower power walking: 65 km on a single battery charge,'' \emph{International Journal of Robotics Research}, 2014.

\bibitem{kim2019highly}
D.~Kim, J.~Di~Carlo, B.~Katz, G.~Bledt, and S.~Kim, ``Highly dynamic quadruped locomotion via whole-body impulse control and model predictive control,'' \emph{arXiv preprint arXiv:1909.06586}, 2019.

\bibitem{ding2021representation}
Y.~Ding, A.~Pandala, C.~Li, Y.-H. Shin, and H.-W. Park, ``Representation-free model predictive control for dynamic motions in quadrupeds,'' \emph{IEEE Transactions on Robotics}, vol.~37, no.~4, pp. 1154--1171, 2021.

\bibitem{neunert2018whole}
M.~Neunert, M.~St{\"a}uble, M.~Giftthaler, C.~D. Bellicoso, J.~Carius, C.~Gehring, M.~Hutter, and J.~Buchli, ``Whole-body nonlinear model predictive control through contacts for quadrupeds,'' \emph{IEEE Robotics and Automation Letters}, vol.~3, no.~3, pp. 1458--1465, 2018.

\bibitem{feng2022genloco}
G.~Feng, H.~Zhang, Z.~Li, X.~B. Peng, B.~Basireddy, L.~Yue, Z.~Song, L.~Yang, Y.~Liu, K.~Sreenath \emph{et~al.}, ``Genloco: Generalized locomotion controllers for quadrupedal robots,'' \emph{arXiv preprint arXiv:2209.05309}, 2022.

\bibitem{chen2023learning}
S.~Chen, B.~Zhang, M.~W. Mueller, A.~Rai, and K.~Sreenath, ``Learning torque control for quadrupedal locomotion,'' in \emph{2023 IEEE-RAS 22nd International Conference on Humanoid Robots (Humanoids)}.\hskip 1em plus 0.5em minus 0.4em\relax IEEE, 2023, pp. 1--8.

\bibitem{jenelten2024dtc}
F.~Jenelten, J.~He, F.~Farshidian, and M.~Hutter, ``Dtc: Deep tracking control,'' \emph{Science Robotics}, vol.~9, no.~86, p. eadh5401, 2024.

\bibitem{koopman1931hamiltonian}
B.~O. Koopman, ``Hamiltonian systems and transformation in hilbert space,'' \emph{Proceedings of the National Academy of Sciences}, vol.~17, no.~5, pp. 315--318, 1931.

\bibitem{williams2015data}
M.~O. Williams, I.~G. Kevrekidis, and C.~W. Rowley, ``A data--driven approximation of the koopman operator: Extending dynamic mode decomposition,'' \emph{Journal of Nonlinear Science}, vol.~25, pp. 1307--1346, 2015.

\bibitem{narayanan2023se}
S.~S. Narayanan, D.~Tellez-Castro, S.~Sutavani, and U.~Vaidya, ``Se (3) koopman-mpc: Data-driven learning and control of quadrotor uavs,'' \emph{IFAC-PapersOnLine}, vol.~56, no.~3, pp. 607--612, 2023.

\bibitem{rahmani2024enhanced}
M.~Rahmani and S.~Redkar, ``Enhanced koopman operator-based robust data-driven control for 3 degree of freedom autonomous underwater vehicles: A novel approach,'' \emph{Ocean Engineering}, vol. 307, p. 118227, 2024.

\bibitem{kim2001model}
B.~Kim, D.~Necsulescu, and J.~Sasiadek, ``Model predictive control of an autonomous vehicle,'' in \emph{2001 IEEE/ASME International Conference on Advanced Intelligent Mechatronics. Proceedings (Cat. No. 01TH8556)}, vol.~2.\hskip 1em plus 0.5em minus 0.4em\relax IEEE, 2001, pp. 1279--1284.

\bibitem{shi2021acd}
L.~Shi and K.~Karydis, ``Acd-edmd: Analytical construction for dictionaries of lifting functions in koopman operator-based nonlinear robotic systems,'' \emph{IEEE Robotics and Automation Letters}, vol.~7, no.~2, pp. 906--913, 2021.

\bibitem{bruder2020data}
D.~Bruder, X.~Fu, R.~B. Gillespie, C.~D. Remy, and R.~Vasudevan, ``Data-driven control of soft robots using koopman operator theory,'' \emph{IEEE Transactions on Robotics}, vol.~37, no.~3, pp. 948--961, 2020.

\bibitem{di2018dynamic}
J.~Di~Carlo, P.~M. Wensing, B.~Katz, G.~Bledt, and S.~Kim, ``Dynamic locomotion in the mit cheetah 3 through convex model-predictive control,'' in \emph{2018 IEEE/RSJ international conference on intelligent robots and systems (IROS)}.\hskip 1em plus 0.5em minus 0.4em\relax IEEE, 2018, pp. 1--9.

\bibitem{li2017extended}
Q.~Li, F.~Dietrich, E.~M. Bollt, and I.~G. Kevrekidis, ``Extended dynamic mode decomposition with dictionary learning: A data-driven adaptive spectral decomposition of the koopman operator,'' \emph{Chaos: An Interdisciplinary Journal of Nonlinear Science}, vol.~27, no.~10, 2017.

\bibitem{todorov2012mujoco}
E.~Todorov, T.~Erez, and Y.~Tassa, ``Mujoco: A physics engine for model-based control,'' in \emph{2012 IEEE/RSJ international conference on intelligent robots and systems}.\hskip 1em plus 0.5em minus 0.4em\relax IEEE, 2012, pp. 5026--5033.

\bibitem{pandala2019qpswift}
A.~G. Pandala, Y.~Ding, and H.-W. Park, ``qpswift: A real-time sparse quadratic program solver for robotic applications,'' \emph{IEEE Robotics and Automation Letters}, vol.~4, no.~4, pp. 3355--3362, 2019.

\bibitem{jovic2015identification}
J.~Jovic, F.~Philipp, A.~Escande, K.~Ayusawa, E.~Yoshida, A.~Kheddar, and G.~Venture, ``Identification of dynamics of humanoids: Systematic exciting motion generation,'' in \emph{2015 IEEE/RSJ International Conference on Intelligent Robots and Systems (IROS)}.\hskip 1em plus 0.5em minus 0.4em\relax IEEE, 2015, pp. 2173--2179.

\bibitem{harry2019dual}
H.~Harry~Asada and F.~E. Sotiropoulos, ``Dual faceted linearization of nonlinear dynamical systems based on physical modeling theory,'' \emph{Journal of Dynamic Systems, Measurement, and Control}, vol. 141, no.~2, p. 021002, 2019.

\bibitem{bruder2021advantages}
D.~Bruder, X.~Fu, and R.~Vasudevan, ``Advantages of bilinear koopman realizations for the modeling and control of systems with unknown dynamics,'' \emph{IEEE Robotics and Automation Letters}, vol.~6, no.~3, pp. 4369--4376, 2021.

\bibitem{folkestad2021koopman}
C.~Folkestad and J.~W. Burdick, ``Koopman nmpc: Koopman-based learning and nonlinear model predictive control of control-affine systems,'' in \emph{2021 IEEE International Conference on Robotics and Automation (ICRA)}.\hskip 1em plus 0.5em minus 0.4em\relax IEEE, 2021, pp. 7350--7356.

\bibitem{yu2022autonomous}
S.~Yu, C.~Shen, and T.~Ersal, ``Autonomous driving using linear model predictive control with a koopman operator based bilinear vehicle model,'' \emph{IFAC-PapersOnLine}, vol.~55, no.~24, pp. 254--259, 2022.

\end{thebibliography}


\end{document}